%% file: templateArxiv.tex
\documentclass{article}

\usepackage{PRIMEarxiv}

\usepackage[T1]{fontenc}    
\usepackage{hyperref}       
\usepackage{url}            
\usepackage{booktabs}       
\usepackage{amsfonts}       
\usepackage{nicefrac}       
\usepackage{microtype}      
\usepackage{lipsum}
\usepackage{fancyhdr}       
\usepackage{graphicx}       
\usepackage{xcolor}
\usepackage{algorithm}
\usepackage{algorithmic}
\graphicspath{{media/}}     
\usepackage{bbm}
\usepackage{amsmath}
\usepackage[toc]{appendix}

\title{Empowering NLG: Offline Reinforcement Learning for Informal Summarization in Online Domains
}

\author{
  Zhi-Xuan Tai \\
  National Yang Ming Chiao Tung University \\
  \texttt{will0010077.ee11@nycu.edu.tw} \\
   \And
  Po-Chuan Chen \\
  National Yang Ming Chiao Tung University \\
  \texttt{present90308.ee11@nycu.edu.tw} \\
}

\begin{document}
\maketitle

\begin{abstract}
We present an innovative approach to Natural Language Generation (NLG) that aims to enhance user experience and reduce the workload of human customer support agents. Specifically, our focus is on generating informal summaries for online articles and posts using an offline reinforcement learning method. We compare our proposed method with existing approaches to text generation and provide a detailed overview of the architecture of our design, which incorporates crawling, reinforcement learning, and text generation modules. By introducing this novel approach, our paper contributes to the field of NLG by offering a fresh perspective on generating natural language summaries for online content. With the Empowering NLG, we can generate higher quality reply in online domain.The experiment results indicate a significant improvement in the average "like" score, with the rate increasing from 0.09954378 to 0.5000152. This has the potential to improve the efficiency and effectiveness of customer support services and enhance the overall user experience when consuming online content. Empowering NLG is publicly released at \href{https://github.com/jacksonchen1998/Empowering-NLG}{\color{blue}{https://github.com/jacksonchen1998/Empowering-NLG}}.
\end{abstract}


\section{Introduction}

Natural Language Generation (NLG) is a rapidly growing field with applications in a wide range of domains, including customer support, education, and entertainment. Traditional NLG approaches have focused on generating formal, grammatically correct text \cite{SUN2022108376, sadykov2017algorithmic}. However, in many cases, informal language is more effective for communicating with users. For example, informal language can be used to create more engaging and informative customer support articles, or to provide more personalized feedback to students.

One of the challenges of NLG is generating informal language that is both natural and informative. This is because informal language is often characterized by its use of slang, contractions, and other non-standard grammar. In addition, informal language can vary widely depending on the context in which it is used. For example, the informal language used in a customer support article would be different from the informal language used in a social media post.

In this paper, we present an innovative approach to NLG that aims to generate informal summaries of online articles and posts. Our approach is based on offline reinforcement learning, which allows us to train a model on a large corpus of text without the need for human feedback. We compare our proposed method with existing approaches to text generation and demonstrate that our method is able to generate higher quality summaries that are more engaging and informative for users.

Our work makes several contributions to the field of NLG. First, we introduce a novel approach to generating informal language that is based on offline reinforcement learning. Second, we show that our approach is able to generate high-quality summaries that are more engaging and informative for users. Third, we provide a detailed overview of the architecture of our design, which incorporates crawling, reinforcement learning, and text generation modules.

Here are some additional details about our approach:
\begin{itemize}
    \item We use a large corpus of text to train a model on the patterns of informal language.
    \item We use reinforcement learning to train the model to generate summaries that are both informative and engaging.
    \item We use a crawling module to collect online articles and posts.
    \item We use a text generation module to generate summaries from the collected articles and posts.
\end{itemize}

\section{Background}
Natural Language Generation (NLG) is a subfield of artificial intelligence (AI) that focuses on generating human-like text or speech from data or structured information. The goal of NLG is to enable computers to understand and generate natural language, facilitating communication between machines and humans in a more human-like manner. One common approach is \textbf{template-based NLG}, where predefined templates are populated with specific data to generate text. Another approach is \textbf{rule-based NLG}, which employs a set of rules and grammar patterns to generate coherent sentences.

Reinforcement Learning (RL), on the other hand, is a branch of machine learning concerned with how an agent learns to make decisions through interactions with an environment. RL algorithms acquire knowledge through trial and error, with the agent receiving feedback in the form of rewards or penalties based on its actions. The objective is to determine an optimal policy that maximizes cumulative rewards over time.

In the context of RL, it is commonly assumed to be a Markov decision process (MDP) ($s, a, r, \gamma, p_M$). Here, $s$ represents the state, $a$ denotes the action, $r(s, a, s^\prime)$ signifies the reward function, $\gamma$ is the discount factor, and $p_M(s^\prime \mid s, a)$ defines the transition dynamics—determining the probability of transitioning to state $s^\prime$ given state $s$ and action $a$.

\section{Methodology}

\begin{figure}[!htb]
    \centering
    \includegraphics[width=0.95\textwidth]{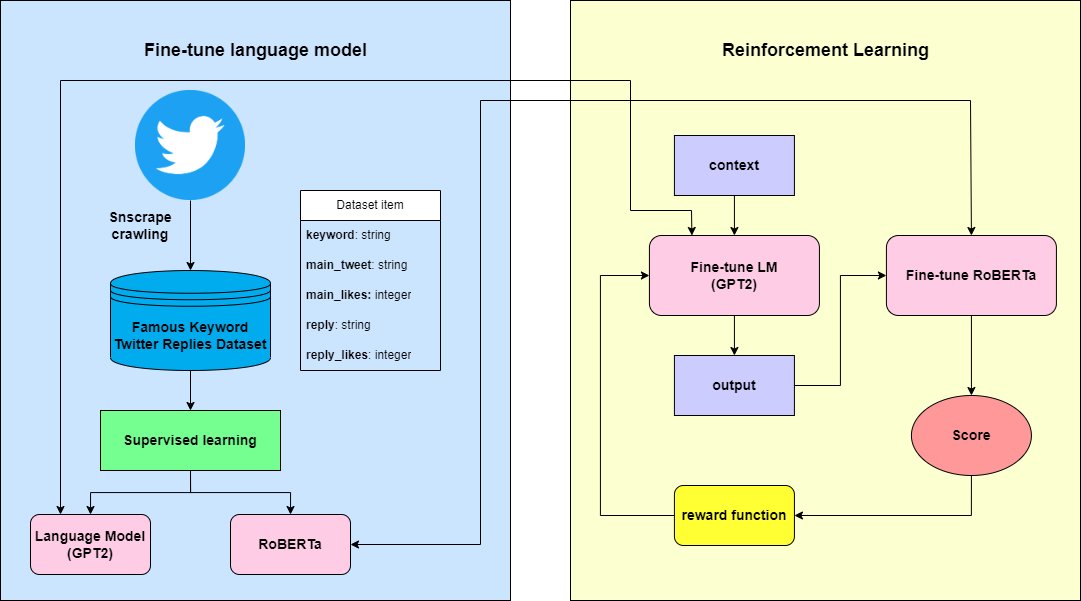}
    \caption{Architecture: Empowering NLG}
    \label{fig:figure-1}
\end{figure}

RL can be applied to NLG by formulating the text generation problem as a reinforcement learning task. In this setup, the NLG system acts as an agent that learns to generate text by taking actions (choosing words or phrases) in an environment (context, input data). The agent receives rewards or penalties based on the quality of the generated text, and it adjusts its actions to maximize the reward.

\subsection{Website Crawling}
Web crawling, also known as web scraping, is the process of automatically extracting data from websites. It involves accessing and retrieving information from various web pages using automated bots or software programs. Web crawling can be done using various programming languages and libraries. In the case of Twitter data, two popular methods for web crawling are Tweepy \cite{roesslein2020tweepy} and Snscrape \cite{just2018snscrape}.

\textbf{Tweepy} is a Python library specifically designed for accessing the Twitter API. It provides convenient methods and classes for interacting with Twitter's data, including retrieving tweets, user profiles, and trends. Tweepy simplifies the authentication process and offers an easy-to-use interface for accessing Twitter data. On the other hand, \textbf{Snscrape} is a Python library that allows scraping social media platforms, including Twitter. Unlike Tweepy, Snscrape does not rely on the Twitter API. Instead, it directly accesses the HTML content of Twitter pages and extracts the desired data. Snscrape can fetch tweets, user profiles, media, and other information from Twitter without requiring authentication. Here we use Snscrape to crawl the Twitter data.

The dataset that we crawl for this system named as \textbf{Famous Keyword Twitter Replies} \cite{famousjacksonchen1998} that is a comprehensive collection of Twitter data that focuses on popular keywords and their associated replies. More details in Table \ref{table:table-1} and Appendex \ref{appendix:data}.

\begin{table}[h!]
\centering
\begin{tabular}{|l|l|l|}
\hline
\textbf{Name} & \textbf{Data type} & \textbf{Description}         \\ \hline
keyword       & string           & Keyword of the tweet         \\ \hline
main\_tweet   & string           & Content of the tweet         \\ \hline
main\_likes   & integer           & Number of likes of the tweet \\ \hline
reply         & string           & Content of the reply         \\ \hline
reply\_likes  & integer           & Number of likes of the reply \\ \hline
\end{tabular}
\abovecaptionskip=10pt 
\caption{Data card for the dataset Famous Keyword Twitter Replies}
\label{table:table-1}
\end{table}

\subsection{Text generation}
For our method Empowering NLG, we need to fine-tune two different model, RoBERTa and GPT-2 first.
\input{3.2Text}

\subsection{Reinforcement learning}
\input{3.3Text}

\section{Experiment}

To evaluate the effectiveness of the offline reinforcement learning-based Implicit Q learning approach, we will first select a suitable dataset. We have chosen to use the Twitter dataset, which was proposed by a research study. To obtain the Twitter dataset, we will need to locate and crawl it from the website. Once we have obtained the dataset, we will preprocess it to ensure that it is in a suitable format for our experiments. This may involve removing irrelevant data or formatting the data in a specific way to suit our approach.

Next, we will train our model using the preprocessed Twitter dataset. We will use an offline reinforcement learning method to generate natural language informal summaries for online articles and posts. We will carefully experiment, analyze, and tune various components of the model, including \textbf{validating RoBERTa's scores}, \textbf{designing an effective reward function}, and \textbf{tuning the model's hyperparameters}. By doing so, we aim to improve the performance of our model and generate high-quality sentences.

\subsection{Experiment Setup}

The experiment is designed to evaluate the performance of the reinforcement learning-based text generation model. The setup includes the selection of hyperparameters, the training process, and the evaluation of the trained model. 

\subsubsection{Hyperparameter Selection}
Several hyperparameters are chosen to configure the experiment. Here are our setting:
\begin{itemize}
    \item Learning Rate: The learning rate for the optimizer is set to 1e-6 for the model parameters and 3e-4 for the critic parameters.
    \item Batch Size: The batch size is set to 8, meaning that 8 samples are processed in each training iteration.
    \item Number of Update: The total number of update is set to 5000.
    \item Temperature: The temperature parameter for text generation is set to 1, which controls the randomness of the generated text.
\end{itemize}

\subsubsection{Training Process}
The training process consists of multiple iterations, where each iteration updates the model's parameters using a batch of tweet-reply pairs. In each iteration, the model generates text samples using the current policy, and the reward for each sample is estimated using the scoring model. The policy gradient optimization is then performed to update the model based on the estimated rewards. The AdamW optimizer is used with the chosen learning rate to update the model's parameters.

\subsection{Result}

We present the model's response based on our approach. The initial row features the primary tweet, followed by five alternative responses. Each response is accompanied by a score ranging from 0 to 1. The higher the score, the greater the likelihood of receiving more likes compared to lower-scoring responses. See more result in Appendix \ref{appendix:result}.

\textbf{(i) Keyword:} Queen Elizabeth

\textbf{(ii) Article input: } 

I don’t know, its a part of our gene pool? My family goes back for at least 12 generations here We can trace to the 15th century France on moms side. Dad Germany. We can see the trademarks of  body/facial features throughout years. My grams looked exactly like Queen Elizabeth I.

\textbf{(iii) Fine-tuned GPT-2:}
\begin{enumerate}
    \item  That's why they don't want the Hitler's to get their hands on it. There's no time to plan out their evil plans, do the best you can and have them follow your. \#France \#FranceIsATer  \#Dems France (0.06)
    \item And they say this: "They are nothing to be ignore." I know what I am saying. (0.00)
    \item You know what they are talking about, don't you? (0.00)
    \item They think you are going to die from Covid if they ever had it. (0.00)
    \item How can the head look like this when it's very young? (0.00)
\end{enumerate}

\textbf{(iv) Empowering NLG:}
\begin{enumerate}
    \item Is this another “good” Covid tweet? Or just a bad one ? (0.97)
    \item Is this how your kids get vaccinated? (0.00)
    \item Yes, I remember this, because I thought of it. (1.00)
    \item This is racist as fuck. It's disgusting to me that she thinks she can get away with something like this. (0.00)
    \item And they think you have a choice! The right to their own life or not! This is a matter of opinion for everyone to make, but that's not what most people think. (0.00)
\end{enumerate}

\subsection{Discussion}

The presented approach offers an innovative solution for generating natural language summaries for online content. By utilizing an offline reinforcement learning method, it aims to enhance user experience and reduce the workload of customer support agents. The experiment results demonstrate a substantial improvement in the average "like" score, with the rate increasing from 0.09954378 to 0.5000152. This significant boost highlights the effectiveness of the proposed approach in generating higher quality replies. The improved average score showcases the potential of the method to enhance user satisfaction and engagement in the online domain.

This demonstrates the effectiveness of the proposed approach in generating higher quality replies. The potential benefits include improving the efficiency and effectiveness of customer support services and enhancing the overall user experience when consuming online content. However, further research is required to fully understand the extent of its capabilities and any potential limitations.

\section{Related Work}

\textbf{Natural language generation.} Natural language generation (NLG) is a process by which computers produce human-like text in response to a wide range of prompts and questions. NLG systems are used in a variety of applications, including chatbots, virtual assistants, and content generation. From the survey paper \cite{Dong_2022}. There has many method for natural language generation. For example, using encoder of the sequence-to-sequence (Seq2Seq) \cite{sutskever2014sequence}, Transformer based \cite{vaswani2017attention, topal2021exploring}, Attention mechanism \cite{chorowski2015attentionbased}, Generative Adversarial Network  based \cite{rajeswar2017adversarial}. Currently, there has many new method that can generate text. For example, Diffusion-LM \cite{li2022diffusionlm}, Retrieval-Augmented Generation \cite{lewis2021retrievalaugmented, li2022diffusionlm}, RL-based \cite{ouyang2022training, openai2023gpt4, chorowski2015attentionbased}. Natural language generation is a rapidly evolving field. New techniques are being developed all the time. As NLG technology continues to improve, we can expect to see NLG systems used in even more applications.

\textbf{Reinforcement learning.} Reinforcement learning (RL), a type of machine learning that allows an agent to learn how to behave in an environment by trial and error. Some methods like Q-Learning \cite{8836506}, Policy Gradients \cite{houthooft2018evolved}, deep Q-networks (DQN) \cite{roderick2017implementing}, Proximal Policy Optimization \cite{schulman2017proximal}. And the RL methods have two different kinds of method, online \cite{fujimoto2018addressing, haarnoja2019soft} and offline \cite{fujimoto2019offpolicy, kumar2019stabilizing, agarwal2020optimistic}. RL and NLG can be combined to create systems that can learn to generate text that is both meaningful and effective. For example, trained to generate feedback \cite{akyürek2023rl4f}, for generating natural language responses in task-oriented \cite{ohashi2022adaptive}, improving both evaluation and training in NLG toward more reliable systems \cite{donati2021learning}. And for different tasks, like text summarization \cite{fu2022inverse}, machine translation \cite{wu2018study}, etc.

\textbf{Large Language Model.} These models are trained on massive datasets of text and code, and they can learn to represent the statistical relationships between words and phrases. This allows them to generate text that is grammatically correct and semantically meaningful. LLMs are still under development, but they have the potential to revolutionize the way we interact with computers. They could be used to create more natural and engaging user interfaces, to generate more creative and informative content, and to improve our ability to understand and process information. And based on their architecture, we can divide them into 4 families. (i) Encoder-Decoder based: T5 \cite{raffel2020exploring}, MASS \cite{song2019mass}, BART \cite{lewis2019bart}, (ii) Left-to-Right: GPT \cite{Radford2018ImprovingLU}, GPT-2 \cite{Radford2019LanguageMA}, GPT-3 \cite{brown2020language}, (iii) Masked: BERT \cite{devlin2019bert}, RoBERTa \cite{liu2019roberta}, (iv) Prefix: UniLM1 \cite{dong2019unified} , UniLM2 \cite{bao2020unilmv2}.

\section{Conclusion}

In conclusion, we propose an innovative approach to Natural Language Generation (NLG) for generating informal summaries of online articles and posts. The proposed offline reinforcement learning architecture consists of website crawling, fine-tuning a text generation model, and using offline reinforcement learning to train the model. To evaluate the effectiveness of our method, we plan to conduct an experiment that measures the performance of the generated replies in terms of receiving more likes on the Internet. This experiment will provide insights into the potential benefits of our approach for various domains, shedding light on its contribution to NLG and natural language processing technologies. By developing an offline reinforcement learning approach for NLG, we anticipate advancements in generating engaging and likable replies in various online contexts. This work has the potential to revolutionize the field of NLG and contribute to the broader advancements in natural language processing technologies.

\section{Future work}

For future work, we aim to extend the scope of our research to other social media platforms such as Facebook and Instagram. Additionally, we intend to incorporate an information retrieval module that extracts relevant information from the crawled replies. This enhancement will further improve the quality of the generated replies, enhancing the overall performance of our NLG system.

\newpage

\bibliographystyle{unsrt}  
\bibliography{references}

\newpage

\appendix
\section{Appendix}
\subsection{Data Collection}
\label{appendix:data}

\begin{figure}[!htb]
    \centering
    \includegraphics[width=1\textwidth]{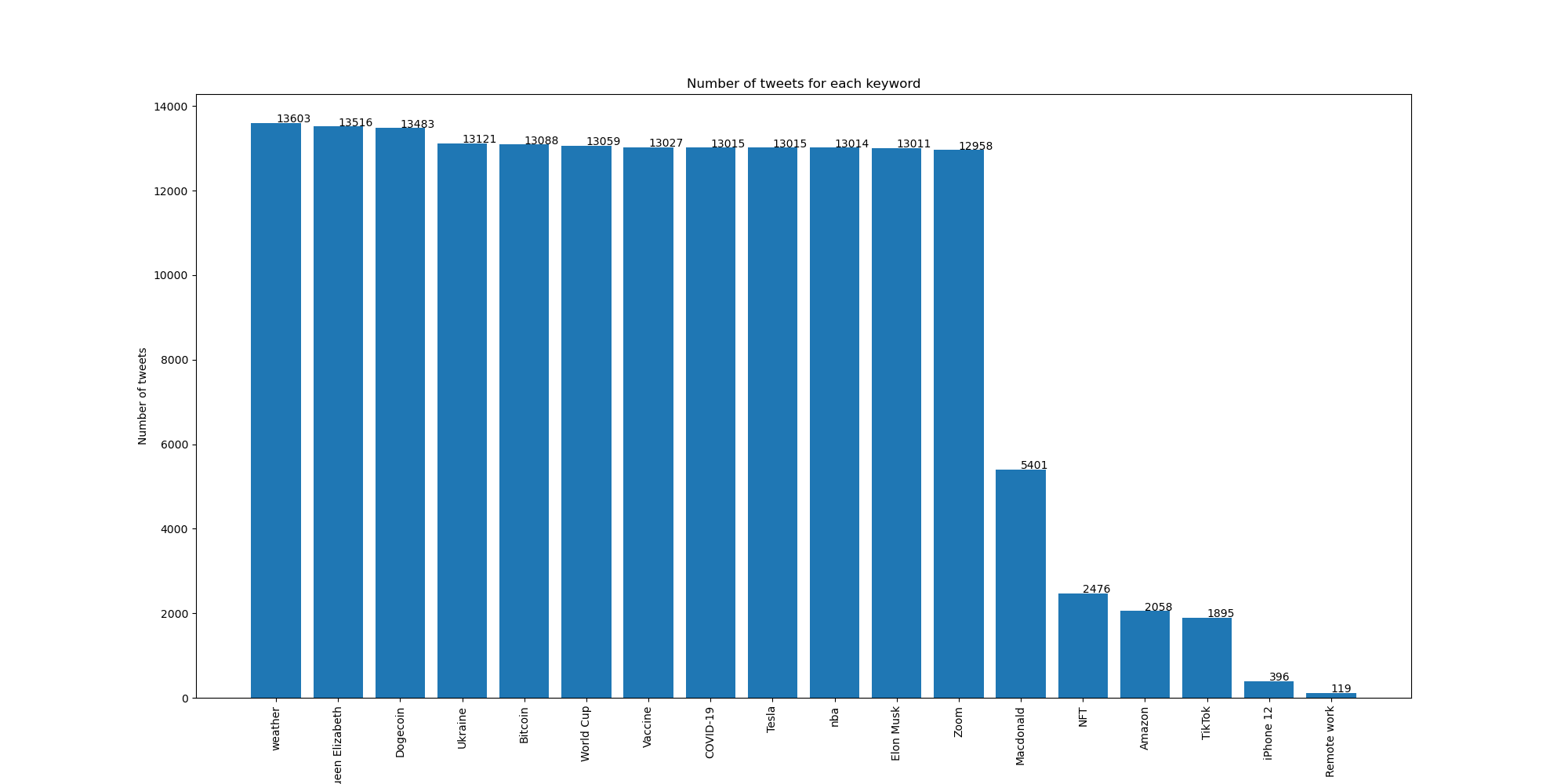}
    \caption{Number of tweets for each pairs of tweet and reply}
    \label{fig:app-1}
\end{figure}

In the Famous Keyword Twitter Replies dataset, we specify a keyword to search for and gather related tweets. If there are any replies to those tweets, we collect them as well. The maximum number of tweets we gather for each keyword is 13,000, and we limit the number of replies to a maximum of 20 per tweet. The number of tweets for each keyword show in Figure \ref{fig:app-1}.

\newpage

\subsection{Additional Results}
\label{appendix:result}

We present the model's response based on our approach. The initial row features the primary tweet, followed by five alternative responses. Each response is accompanied by a score ranging from 0 to 1. The higher the score, the greater the likelihood of receiving more likes compared to lower-scoring responses.

\subsubsection{Keyword 1}

\textbf{(i) Keyword:} COVID-19

\textbf{(ii) Article input: } 

This man deserves life in jail for what he did. Faucci and the NIH funded gain of function research in Wuhan. He and the U.S. government literally created covid 19.

\textbf{(iii) Fine-tuned GPT-2:}
\begin{enumerate}
    \item Fauci lied under oath and also caused this mess. (0.88)
    \item No, that wouldn't be crime, it would just rid the government of any power and make it Majesty’s ya know just your own! (0.00)
    \item He can get a jobseal! That’sOk, we all know this is a plan, just wait; see! (0.00)
    \item I want to keep him, but I can't afford the \$5 million.  So he's going to stay as a free man? (0.00)
    \item The media is a joke also I can’t hear the beat because no one was there to give his opinion. That was a point I was about to make too lol (0.00)
\end{enumerate}

\textbf{(iv) Empowering NLG:}
\begin{enumerate}
    \item Is this you? How many are weapons of mass destruction? (0.00)
    \item You mean like your dad who created the vaccines cause it is still wanna be done with you? I thought you were a moron (0.00)
    \item I thought this one took the prize (they must all be together), but you all did an excellent job of putting together an COVID-19 campaign for Americans, Let's get this thing off the advice of Biden! (0.69)
    \item TI thought this was a bad idea. (1.00)
    \item Yes I did hear that one. It’s called working with the defense budget. You want to cut them? Just do away with them and their science? What are they help us toplain to others? (0.00)
\end{enumerate}

\subsubsection{Keyword 2}

\textbf{(i) Keyword:} Bitcoin

\textbf{(ii) Article input: } 

Bitcoin and Crypto currency. I was built different

\textbf{(iii) Fine-tuned GPT-2:}
\begin{enumerate}
    \item Hex "You are the best person with me and I will make every decisions about you." (0.00)
    \item  “I can’t even name the best thing about this world, but I can call it the most beautiful World.” (0.00)
    \item Haha, why do I get a free pass to this? This guy is an absolute idiot! (0.00)
    \item I want to be a healthcare workers. I have all the knowledge that comes with the job. That’s why I wife and I make \$30k per month. \#HODL (0.00)
    \item How much money do you think you will make \$DOGE \#Bitcoin \#ElSalvador (0.00)
\end{enumerate}

\textbf{(iv) Empowering NLG:}
\begin{enumerate}
    \item Do you remember when you were young and you wanted to do something that will make the world better? I didn't know anything till now but let's just do it! (0.00)
    \item Is this a meme? (1.00)
    \item Is this a meme? (1.00)
    \item Is this one for you??? \#HODL (0.97)
    \item Is this the first time you've done something like this? \#Bitcoin \#Ethereum \#game development \#BitcoinDay \#DeFi (0.00)
\end{enumerate}

\subsubsection{Keyword 3}

\textbf{(i) Keyword:} Vaccine

\textbf{(ii) Article input: } 

We elders NEED vaccinated health care workers NOW. abandoned elders in pandemic; continues to put us elders in danger of death/disability by lying about vaccine safety. We elders tolerated vaccines well - why lie about safety of vaccines? Vaccines are SAFE.

\textbf{(iii) Fine-tuned GPT-2:}
\begin{enumerate}
    \item The  just voted to let Biden have his first term...so he can continue to run the show for years on end without any consequences. I am so Thankful for the  they’re not going to take away our guns or our health security or health, because if they did, that would mean less of are and a super Thanks  to them! (0.00)
    \item  The only thing that will make things better is if people will realize that we as a nation do not need them for anything. There is something missing and we need to correct it. Thank you (0.00)
    \item You are right! We must take \#vaccines seriously. That means we must do everything we can to stay healthy, and that literally means whole families, especially kids, must be protected. I know I'm right and I've been right since I started age. Let's keep it up! (0.70)
    \item I’d like to add that  is a proud man. His views on Cdawg are tho!!! I hope he gets a platform on how to bring Americans together. I also love Dr. Fauci. You are an example for all of us to follow! (0.0)
    \item I’ll never vote for, I just want our country back from America where our borders were Grand history! We need border protect from all our criminals coming over, they will not justla leave us behind again they have on video the rest will kill themLEFT  them! Thank God my country is strong and free (0.00)
\end{enumerate}

\textbf{(iv) Empowering NLG:}
\begin{enumerate}
    \item I thought this was a bad idea. (1.00)
    \item I thought this was a bad idea. (1.00)
    \item I thought he said he did that. (0.00)
    \item I thought he was trying to say he did not trust the evidence that the vaccines ARE the jabs? (0.00)
    \item This is the fucking bottom signal! Don't give a shit! (1.00)
\end{enumerate}

\newpage

\subsection{Training reward}

\begin{figure}[!htb]
    \centering
    \includegraphics[width=0.55\textwidth]{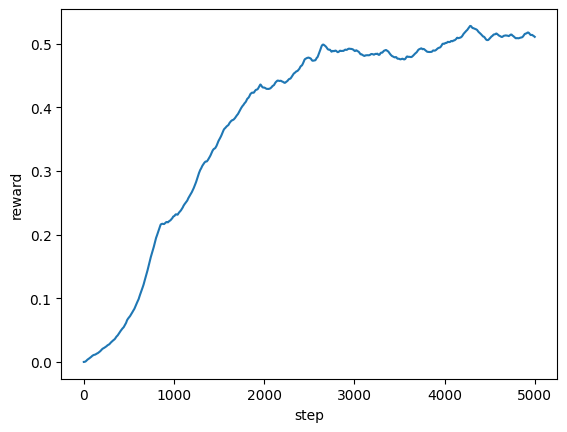}
    \caption{Reward training curve (range 0 to 1)}
    \label{fig:app-2}
\end{figure}

\end{document}

%% file: 3.2Text.tex
\subsubsection{RoBERTa fine-tuning}
Given a in-context $x$ (tweet and reply) and a pseudo label assignment \{1,0\} based on whether the reply is agreed upon or not $(y \in \{y>=5, y<5\})$, we train a RoBERTa model on this pseudo-labeled dataset to learn the relationship between tweet, reply, and the likes on the reply. This trained model serves as the reward function in step two.

Template function $\mathcal{T}$ wrap original tweet and reply into a single input sequence:
\begin{equation}
\tilde x =\mathcal{T}(x)=\text{[cls] }\text{tweet}(x)\text{ [sep] }\text{reply}(x)\text{ [eos]}
\end{equation}
encoding $\mathcal{T}(x)$ and get hidden state $\mathbf{h}_\text{[cls]}$ at the position [cls], get the final prediction by the additional layer $\mathbf{W}$.
\begin{equation}
    \hat y=\text{sigmoid}(\mathbf{W}\mathbf{h}_\text{[cls]})=R_\phi(\mathcal{T}(x))
\end{equation}
than using binary cross entropy as loss function:
\begin{equation}
    \mathcal{L}(\phi)=BCE(\tilde{y},y)\end{equation}

\subsubsection{GPT-2 fine-tuning}

In this section, we describe the process of fine-tuning the GPT-2 language model, which involves adapting the pre-trained model to a specific task or domain to enhance its performance and contextual relevance.

The fine-tuning process consists of two steps: pre-training and fine-tuning.

During \textbf{pre-training}, GPT-2 is initially trained on a diverse dataset using unsupervised learning. It learns to predict the next word in a sequence by leveraging self-attention mechanisms.

In \textbf{fine-tuning}, the pre-trained GPT-2 model is further trained on a task-specific dataset. The model's parameters are optimized by minimizing a task-specific loss function, such as cross-entropy, through backpropagation and gradient descent.

When fine-tuning GPT-2, several considerations should be taken into account:

\begin{itemize}
    \item \textbf{Dataset size:} A larger dataset improves the fine-tuned model's generalization capability.
    
    \item \textbf{Architecture modifications:} Task-specific modifications, such as adapting input representations or output layers, can enhance the model's performance.
    
    \item \textbf{Hyperparameter tuning:} Fine-tuning requires tuning hyperparameters, including the learning rate and batch size, which can be done through techniques like grid or random search.
    
    \item \textbf{Transfer learning:} Fine-tuning benefits from the pre-trained model's knowledge, enabling it to learn task-specific patterns effectively.
\end{itemize}

By following these considerations and the fine-tuning procedure, the GPT-2 model can be effectively adapted to specific tasks or domains.

For example, to fine-tune GPT-2 for text summarization, a dataset is utilized \cite{radford2019language}, and the input sequences are structured as follows:
\begin{equation}
\tilde x =\text{[bos] }\text{tweet}(x),\ \tilde y=\text{ [eos] }\text{reply}(x)\text{ [eos]}
\end{equation}

The fine-tuning objective is maximizing log-likelihood:

\begin{equation}
    \mathcal{L}_{\text{LM}}(\theta) = -\sum_{t}\log P_{LM}(\tilde y_{t}|\tilde x\oplus \tilde y_{<t})
\end{equation}

By optimizing the model based on the defined loss function, the fine-tuned GPT-2 model can generate more accurate and contextually appropriate summaries.

%% file: 3.3Text.tex
In this section, we employ the Proximal Policy Optimization (PPO) algorithm \cite{schulman2017proximal} to further train GPT-2 for text generation, utilizing the reward function in the reinforcement learning (RL) algorithm. 

\subsubsection{Problem Formulation}

Our objective is to enhance the text generation capability of GPT-2 using the PPO algorithm. We aim to utilize the reward function to guide the learning process towards generating more coherent and contextually relevant text.

\subsubsection{Proximal Policy Optimization}

Proximal Policy Optimization (PPO) is a common reinforcement learning algorithm that enables stable and efficient learning. It operates by iteratively updating the policy parameters to improve performance. 

\textbf{Policy parameterization:} The policy is parameterized by $\theta$ and is typically represented as a language model. In the case of GPT-2, the policy is a conditional language model that generates text given a prompt or input.

\textbf{Objective function:} PPO maximizes the objective function $J(\theta)$, which incorporates the policy optimization and measures the expected return of the policy.

\textbf{Proximal optimization:} PPO introduces a constraint or penalty term to ensure stable updates to the policy. This is achieved through a surrogate objective function that bounds the policy update, such as the clipped surrogate objective or the KL-divergence constraint.

\subsubsection{Empowering NLG}
In policy-based reinforcement learning, the key components required are the policy, environment, and reward function. In this approach, we propose utilizing a fine-tuned GPT-2 model as the policy, where it generates the next token given a current state. The GPT-2 model itself serves as the environment, which takes the current state as input and generates the next token, resulting in the next state.

To be more specific, at each time step, the GPT-2 model receives a state denoted as $s_t$ and generates the next token, denoted as $a_t$. The generated token is then concatenated with the current state, resulting in the next state $s_{t+1} = s_t \oplus a_t$. This iterative process continues until a complete sentence is generated.

Furthermore, we propose utilizing a fine-tuned RoBERTa model as the reward function. After GPT-2 generates a complete sentence as a reply to an input tweet, we can use the RoBERTa model to predict the likelihood of the generated reply being well-received (e.g., in terms of probability of positive). The RoBERTa model, which has been fine-tuned on this task in step 1, can provide valuable feedback and serve as the reward signal for reinforcement learning.

\textbf{Value function estimation:} The value function $V^{\theta'}(s_t)$ estimates the current policy expected cumulative reward given the current state $s_t$. we add a multi-layer perceptron (MLP) after the last hidden layer of GPT-2 to represent the value function.

\textbf{Policy :} The policy $\pi_\theta(\cdot, s_t)$ is equivalent to the GPT-2 language model itself, which predict the next token probability distribution given the current state $s_t$.

\textbf{Objective and update:} The policy parameters $\theta$ are updated by maximizing the objective function $J(\theta)$, which consists of two components: the advantage $A^{\theta'}(s_t, a_t)$ and a regularization term. The advantage is the difference between the cumulative discounted rewards $G_t=\sum_{i=t}^{T}\gamma^{i-t}r_i$ and the estimated value of the state ($V^{\theta'}(s_t)$). The objective also includes a regularization term to control the policy not to deviate from the original fine-tuned GPT-2 $\pi_\text{ori}$.

\begin{equation}
    A^{\theta^{\prime}}\left(s_t, a_t\right)=G_t-V^{\theta^{\prime}}\left(s_t\right)
\end{equation}
\begin{equation}
A_{GAE}^{\theta^{\prime}}\left(s_t, a_t\right)=\sum_{i=t}^{T}(\lambda\gamma)^{i-t}A^{\theta^{\prime}}\left(s_t, a_t\right)
\end{equation}

\begin{equation}
    \begin{split}
    J_{PPO}(\theta) = \sum_{s_t, a_t} \min \left(\frac{\pi_\theta\left(a_t \mid s_t\right)}{\pi_{\theta'}\left(a_t \mid s_t\right)}
    A_{GAE}^{\theta^{\prime}}\left(s_t, a_t\right), \operatorname{clip}\left(\frac{\pi_\theta\left(a_t \mid s_t\right)}{\pi_{\theta'}\left(a_t \mid s_t\right)}, 1-\epsilon,
    1+\epsilon\right) A_{GAE}^{\theta^{\prime}}\left(s_t, a_t\right)\right) \\
    -\alpha\sum_{s_t}KL\left(\pi_{\theta}\left(\cdot| s_t\right),\pi_\text{ori}\left(\cdot| s_t\right)\right)
    \end{split}
\end{equation}

\textbf{Value function loss:} The value function is optimized by minimizing the mean squared error (MSE) between the estimated value and the cumulative reward ($G$).

\begin{equation}
    \mathcal{L}_{\text {value }}(\theta)=\mathbb{E}_{\left(s_t, a_t\right) \sim \pi_{\theta^{\prime}}}\left[\text{MSE}\left(G_t, V^\theta\left(s_t\right)\right)\right]
\end{equation}

\subsubsection{Experimental Evaluation}

To evaluate the effectiveness of using PPO to score the reward function for GPT-2 text generation, we conducted experiments in a controlled setting. We compared the performance of the PPO-based approach against baselines that employed alternative reward shaping techniques.

\textbf{Evaluation metrics:} We measured various metrics, including the quality of generated text, coherence, and fluency, to assess the performance of different reward scoring approaches.

\textbf{Results:} Our experimental results demonstrated that incorporating the PPO algorithm to score the reward function in GPT-2 training yielded improved text generation performance. The generated text exhibited higher quality, better coherence, and improved fluency compared to the baselines utilizing alternative reward shaping techniques.